# Can structural correspondences ground real world representational content in Large Language Models?

Iwan Williams (ORCID: 0000-0003-0582-0983)
iwan.r.williams@gmail.com

Centre for Philosophy of AI, University of Copenhagen

**Abstract**:

Large Language Models (LLMs) such as GPT-4 produce compelling responses to a wide range of prompts. But their representational capacities are uncertain. Many LLMs have no direct contact with extra-linguistic reality: their inputs, outputs and training data consist solely of text, raising the questions (1) can LLMs represent *anything* and (2) if so, *what*? In this paper, I explore what it would take to answer these questions according to a *structural-correspondence* based account of representation, and make an initial survey of this evidence. I argue that the mere existence of structural correspondences between LLMs and worldly entities is insufficient to ground representation of those entities. However, if these structural correspondences play an appropriate role—they are exploited in a way that explains successful task performance—then they could ground real world contents. This requires overcoming a challenge: the text-boundedness of LLMs appears, on the face of it, to prevent them engaging in the right sorts of tasks.

1. Introduction

Large Language Models (LLMs) are artificial systems designed to model, process and/or generate, natural language. Historically, these systems included purely statistical models, but modern LLMs are deep artificial neural networks trained via machine learning. Once trained, an LLM may be implemented for various purposes, such as in chatbots and personal assistants, or for translation, sentiment analysis and document review.

The indisputably impressive *performance* of LLMs on a wide variety of task raises pressing questions about their *capacities*, and the mechanisms underlying those capacities. For instance, authors have grappled with the questions of whether LLMs understand language (Bender & Koller, 2020; Mitchell & Krakauer, 2022) whether they possess concepts (Butlin, 2023) or to what extent they possess a theory of mind (Kosinski, 2024; Ullman 2023).

This paper focuses on the *representational* capacities of LLMs. Do LLMs rely on representations? If so, *what* do those representations represent? Much research in AI—for instance, studies using probing classifiers (Belinkov, 2022), and methods for "editing" models' representations (Hernandez et al., 2024; Meng et al., 2022)—assumes that a representational lens is appropriate. But a key question is whether LLMs can represent real world entities, or only "shallow" linguistic contents that don't reach into extra-linguistic reality (Butlin, 2021; Coelho Mollo & Millière, 2023; Yildirim & Paul, 2024). Researchers posit representations of tokens, words, sentences (Pavlick, 2023), distributions over words (Geva et al. 2021), and other syntactic features like roles and parts of speech (Rogers et al., 2020). But some occasionally posit representations of real world, extra-linguistic contents like time and space (Gurnee & Tegmark, 2024), relations between colours (Abdou et al., 2021) and facts of various kinds (Hernandez et al., 2024).

Much of this research has not been grounded in the philosophical literature on representation, but there have been some recent attempts to adapt and apply philosophical accounts of representation in biological organisms to evaluate the representational capacities of LLMs (Butlin, 2021; Harding, 2023; Coelho Mollo & Millière, 2023). These discussions have primarily explored accounts of representation based on informational relations. But an alternative account of representation in the literature, which has so far not received sustained investigation as applied to LLMs, is a *structural-correspondence* based account.

Structural correspondence accounts are a popular family of accounts of mental representation in philosophy of mind and cognitive science (Cummins, 1996; Gładziejewski & Miłkowski, 2017; Lee, 2019; O'Brien & Opie, 2004; Ramsey, 2007; Shea, 2014, 2018; Swoyer, 1991). In addition to its popularity as an account of mental representation in general, there are particular reasons to think that this approach might be fruitfully applied to LLMs. For example, evidence of structural correspondences between the internal states of LLMs and various real world domains has been used as a basis for ascribing representations of world-involving content to LLMs (Søgaard, 2022; 2023).[1] Thus, some seem to accept, at least implicitly, a view on which structural correspondences can play a role in grounding representation.

Drawing on the philosophical literature, I will caution against being too quick to infer representation of world-involving contents from the discovery of a structural correspondence. After clarifying what it takes for a structural correspondence to play a genuinely content-grounding role, I will argue that it is a live empirical possibility that text-based LLMs represent world-involving contents. Before making that case, I will lay out the primary challenge for ascribing world-involving content to LLMs, namely their text-bound nature.

## 2. The text-boundedness challenge

A key challenge for any attempt to ascribe world-involving contents to LLMs is that many of them, including those implemented in simple chatbots, are *text-bound*, having no direct contact (via perception or action) with extra-linguistic reality: their inputs, outputs and training (aside from human feedback signals) consist solely of text. Thus, on the face of it, it is hard to see how they could represent anything other than text.

In their influential critique, Bender & Koller (2020) introduce a thought experiment—similar in many respects to Searle's (1980) "Chinese Room"—with the goal of highlighting that a text-only LLM could at most represent *formal* properties of language, never *semantic* properties. Bender & Koller ask us to imagine two English speakers, A and B, trapped on separate desert islands, who discover telegraphs connected by underwater cables, allowing them to communicate with each other. A super-intelligent octopus chances upon the underwater pipe and intercepts it, listening in to the telegraph signals passing through it. While the octopus doesn't understand English (at least at first), it is skilled at detecting statistical patterns. It starts to notice regularities in the exchange—certain signals occur in similar contexts—and builds up a vast memory of them. One day, the octopus breaks the pipe and inserts itself into the conversation. Now, the messages sent by A no longer get sent to B, rather, they get sent to the octopus, who, based on its knowledge of the most likely responses to certain patterns of signals, is able to send messages back which are similar to the sorts of thing B would say in response.

---

[1] Yildirim & Paul (2024) develop the hypothesis in terms of "world-models" which they define as "structure-preserving, behaviorally efficacious representations of the entities, relations, and processes in the real world" (p. 3).

The intuition elicited by this thought experiment, argue Bender & Koller, is that the octopus, despite its ability to generate signals which appear to be sensible responses to the recipient's messages, has no understanding of the *meanings* of the messages it sends. After all, while A and B may use telegraph messages to refer to various objects, the octopus has no experience with those objects and, more importantly, no way of connecting those objects to those messages. For example, a telegraph mentioning "coconuts" is used by A to refer to coconuts. But the octopus—so the intuition goes—has no way of representing coconuts. Text-bound LLMs, Bender & Koller argue, are like the octopus in this thought experiment. Having access only to formal symbols and their statistical relations, they cannot represent the real world entities that we use those symbols to represent.[2]

The octopus thought experiment is most forceful (at least on the face of it) if we assume that causal or informational relations to the external world are the only means of grounding world-involving representational contents. Text-bound LLMs don't bear the right sort of causal/informational relations to extralinguistic entities and so, the argument goes, they cannot represent extra-linguistic reality.[3] Some authors have pointed to structural correspondences precisely because they seem to provide a way of responding to Bender and Koller's thought experiment (Søgaard, 2023). However, below, I'll show that those who want to appeal to structural representation to ground world-involving contents must also reckon with a version of the text-boundedness worry.

---

[2] Strictly speaking, Bender and Koller are interested in the question of whether LLMs understand natural language expressions, rather than our question—whether they have internal representations of real world (non-linguistic) contents. But the thought experiment raises analogous challenges for the representation question.

### 3. Structural correspondence accounts

Structural correspondence accounts of mental representation hold that representations are grounded (at least partly) in a morphism, resemblance, or mirroring between a set of internal states or vehicles (e.g. in an organism's brain) and a set of external entities (e.g. out in the world). The relevant notion of structural correspondence is a relation-preserving map: there's a way of assigning internal vehicles to external entities such that a pattern of relations among the former preserves a pattern of relations among the latter.[4] According to such accounts, what makes a relational structure $R$ a representation of another relational structure $S$ is (at least in part, and for at least some representations) the fact that $R$ bears a structural correspondence to $S$. The "for at least some representations" clause in this claim allows for stronger views (those which hold that structural correspondence is the *only* viable representation-grounding relation) and weaker views (those which hold that other features or relations, such as correlation, are also capable of grounding content).[5] The "at least in part" clause captures the fact that theorists rarely take the existence of a structural correspondence to be *sufficient* for making something a representation or fixing its content. Structural correspondences are ubiquitous and, thus, an account which appealed only to structural correspondences would overgeneralise hopelessly, in effect classifying any structure as a representation, and attributing radically indeterminate contents to them.[6]

---

[3] An anonymous reviewer suggests that an LLM might bear indirect causal-informational relations to real world objects: its internal states might correlate with occurrences of words (e.g. "coconut") which in turn correlate with occurrences of the entities that those words refer to (the presence of a coconut in the vicinity of the speaker). It is beyond the scope of this paper to fully evaluate the prospects of causal-informational grounding of content in LLMs. But a significant hurdle to developing this proposal is the fact that occurrences of words are often radically decoupled from occurrences of their referents: we often talk about absent, past, future, possible, false, abstract and fictional scenarios.

[4] See O'Brien and Opie (2004) and Shea (2018, p. 117) for attempts to formalise the notion of structural correspondence.

[5] To mark this distinction, Nirshberg (2023) distinguishes between "strong structuralism" and "moderate structuralism".

There is some disagreement between theorists as to what additional conditions must be met for a structural correspondence to play a genuinely content grounding role. However, there is broad agreement on the general shape of those conditions: to a first approximation, the structural correspondence needs to be *used* or *exploited* by the system (or organism) in a way that explains the system's successful performance of some task or capacity.[7] Lee and Calder summarise this general consensus (here "S-representation" refers to the kind of structural-correspondence based representation we've been discussing):

> S-representations explain when the following conditions apply: (1) a system undertakes some task (e.g., navigating to a target location); (2) the outcome of that task depends on a mechanism with a component that structurally resembles task-relevant items (e.g., a cognitive map); (3) success depends on the degree of structural similarity between the mechanism component and those items (e.g., the topographical features of a rat's local environment). (Lee & Calder 2023)

Perhaps the most famous example of such a representation in the literature is the system of place-cells in mammalian hippocampi (O'Keefe, & Nadel 1978; O'Keefe & Burgess 1996). Each place-cell is preferentially active when the animal is in a certain known location and the patterns of coactivation between place-cells in a rats' hippocampus structurally correspond to spatial relations between the places which activate those place cells. For example, there is a way of mapping place-cells to locations, such that if two places are adjacent, the corresponding place-cells will have strongly excitatory connections (and thus tend to have highly correlated activity). The structural correspondence between place-cells and locations has been hypothesised to play a role in grounding the representational content of those place-cells and their co-activation relations (Shea 2018). For instance, several models of mammalian route-planning propose mechanisms whereby the place cell network is taken "offline" and the structural correspondence between co-activation relations and spatial proximity is exploited to identify the shortest route to some goal location (Corneil & Gerstner, 2015; Khajeh-Alijani, Urbanczik, & Senn, 2015; Reid & Staddon, 1998).

---

[6] For discussions of this challenge see (Godfrey-Smith, 1996, pp. 184–187; Isaac, 2013; Shea, 2013). A concern over the triviality of isomorphism has arisen in various debates, at least as early as Newman (1928).

[7] Exploitable relations in the context of representation are discussed in (Godfrey-Smith, 2006). Many advocates of structural correspondence based accounts of representation grounding have made appeals to exploitability (Shea 2018; Gładziejewski & Miłkowski, 2017; Isaac, 2013; Williams & Colling, 2018). Others appeal to related notions like interpretation (O'Brien & Opie, 2004) and action guidance (Gładziejewski, 2015, 2016). See (Lee, 2019; Nirshberg, 2023) for discussion.

This example involves a structural correspondence with physical space, but structural correspondences (and representations involving them) could involve relational structures of many different kinds—examples from cognitive science include representations of social relations (Park, Miller, & Boorman, 2021) and causal relations (Tenenbaum, Kemp, Griffiths, & Goodman, 2011). And, crucially, for a structural correspondence to bear between an internal (e.g. neural) structure and an external structure, the relation comprising the internal structure need not be *the very same relation* as that comprising the external structure (e.g. spatial relations corresponding to spatial relations)—structural correspondence is an abstract kind of resemblance, which involves a preservation of a *pattern* of relations, rather than requiring duplication of the very same relation type.

In what follows, I will not directly argue for the viability of structural correspondence as an account of mental representation (though my exposition will hopefully go some way towards motivating such accounts). Nor will I rehearse arguments for the legitimacy and value of representational explanation in general. I will assume, for the sake of argument, that a well-developed structural correspondence based theory of representation (e.g. one that builds in a robust exploitation criterion) can be used to establish the existence and content of representations in at least some cases. The scope of this paper will be to apply the structural correspondence approach to LLMs, and see whether we find evidence of representations with world-involving contents.[8]

## 4. Applying the account to LLMs

### 4.1. Evidence of structural correspondences between LLMs and worldly domains

A common tool in the analysis of artificial neural networks, including LLMs, is to conceptualise the possible activity of a given layer of a network in terms of a *state space* (sometimes also referred to as an "embedding space"). The activity of each individual unit defines a dimension or axis of that space. (So, a layer with n units can be characterised in terms of an n-dimensional state space). The pattern of activation (the activation vector) across a layer at a particular time can be characterised as a *point* in that state space, that is, a position along each of the n dimensions. There are many different relations between points or regions of a state space, e.g. two points might be *closer* to each other than two others. Or a particular region might be a *proper part* of another region.

Tools such as Representational Similarity Analysis (Kriegeskorte, Mur, & Bandettini, 2008) and "probing" methods (Belinkov 2022), have revealed structural correspondences between, on the one hand, the various internal state spaces of LLMs and, on the other hand, relational structures outside the network. I will consider four examples of structural correspondences from the empirical literature. Firstly, it has long been observed, even in simpler language models, that, for single word inputs, words with similar meanings are assigned to nearby points in activation space.[9] For example, the vectors generated for the inputs "country", "town" and "city" might be clustered close together in this state space, while the vectors generated for "income", "payment" and "costs" form another cluster, some distance from the first cluster. Thus, there seems to be a structural correspondence between, on the one hand, proximity in LLM activation state space and, on the other, similarity at the level of word meaning. Beyond mere similarity, certain vector offsets between points in LLMs' internal activation spaces, correspond to specific relations between those words. For example in an early finding with a recurrent neural network based language model, Mikolov et al. (2013) showed that via vector arithmetic one could roughly recover the vector corresponding to "queen" by subtracting the vector for "man" from the vector for "king" then adding the vector for "woman".

---

[8] I will also not argue against alternative accounts of representation, such as information-theoretic approaches. There may be a plurality of legitimate representation-grounding conditions, so by showing that LLMs can or cannot represent real world contents on a structural correspondence based approach I will not have shown that they can or cannot on some alternative account. Those who remain sceptical of structural correspondence as an approach to representation may nonetheless find the discussion useful in clarifying the consequences of these accounts when applied to LLMs.
[9] Activation spaces for words in LLMs are often referred to as "embeddings".

In a more recent study, Abdou and colleagues (2021) investigated patterns of activation in LLMs (BERT, RoBERTa and ELECTRA) trained on natural language data, and discovered structural correspondences between the activation spaces in these models and perceptual colour spaces. They compared the activation space induced by prompting models with colour terms ("colour term embedding space"), and a space encoding colours in a 3D coordinate scheme ("CIELAB space"). Using two analysis methods, they found modest but statistically significant structural correspondences between the two spaces. For example, when they compared the patterns of activation induced after giving the LLMs a prompt mentioning "violet", a prompt mentioning "turquoise", a prompt mentioning "olive" and so on, the profile of similarities and differences between those patterns of activities mirror (to some degree) the profile of similarities and differences between the colours themselves (as encoded in the CIELAB colour coordinate scheme) (see Figure 1). Intuitively, terms for similar colours induce similar patterns of activation (patterns that are closer in colour term embedding space).

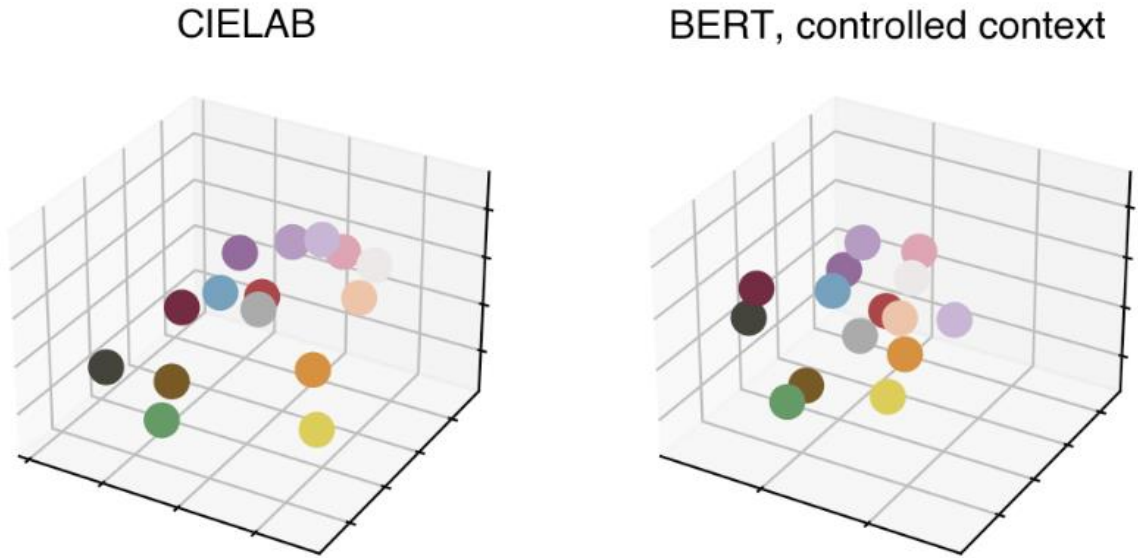

Figure 1: Left: Colour orientation in 3d CIELAB space. Right: linear mapping from BERT colour term embeddings to the CIELAB space. (from Abdou et al. 2021)

In a different study, Gurnee & Tegmark (2023) identified structural correspondences between LLMs' activation spaces and temporal and spatial structures. They trained LLMs (LLaMa-2, Pythia) on a tailored dataset which included text describing dates and locations of world events and landmarks. The models, once trained, were able to produce fairly accurate responses to questions about times and locations (including generalising to locations and times absent in the training data). Additionally, using a linear probe (a separate classification network trained to predict properties of inputs from internal patterns of activity in the LLMs), the experimenters could decode locations and times from the activity of the networks at certain layers. Importantly for our purposes, activity seemed to be organised along particular dimensions, such that when a prompt mentioned a particular event, the activity along that dimension would be intermediate between the activity caused by prompts mentioning earlier events and later events. In other words, there seemed to be a structural correspondence between the activity along certain dimensions, and the temporal ordering of events (a similar effect was seen for spatial relations).

What should we make of these structural correspondences? Sometimes, it appears as if researchers take the existence of such a correspondence to be sufficient to ground representation, or they are at least not explicit about what further conditions they take to be satisfied. However, this inference from structural correspondence to representation would be too quick. As noted above, structural correspondences are cheap, and so the mere existence of a structural correspondence cannot ground representation, on pain of trivialising the notion of representation. To justify the claim that these observed structural correspondences play a role in grounding representation of real world contents, we need to determine whether these correspondences are *exploited*.

To sharpen up the challenge, notice that when we observe a correspondence between internal states of an LLM and some worldly structure, there are at least two alternative explanations of this correspondence, which compete with the claim that the LLM represents the worldly structure. Firstly, given the liberality of the relation of structural correspondence (discussed in the previous section) the correspondences between internal states and external states could be purely coincidental, offering no explanatory insight into the LLM's behaviour. Secondly, and more subtly, a correspondence with a structure *S* could be an artefact of the fact that *S* structurally corresponds to some further structure *T*, where the correspondence with *T* more directly explains the system's behaviour.

This latter possibility is important given the current dialectic: our question is whether text-bound LLMs can represent *real world* contents, as opposed to merely linguistic objects, properties, and relations. Suppose we find evidence of a structural correspondence between internal states of an LLM and a structure in the real world (such as similarity relations between colours). It may be that what's really going on is that the LLM is exploiting a structural correspondence between its internal states and the statistics of natural language colour terms (e.g. the patterns of use of terms like "turquoise", and "navy", including which other tokens they tend to occur in the context of). Thus, the observed structural correspondence to real world colour space may not itself be exploited by the system, but could simply be a side effect of the fact that the statistics of natural language themselves structurally correspond to that real world structure.[10] It is thus crucial to establish that a correspondence with a real world structure is exploited, if it is to ground representational contents in LLMs. In the remainder of this paper, I will evaluate the evidence for these real world structural correspondences being exploited.

### 4.2. Evidence of exploited structural correspondences

It will be useful to tease apart two aspects of the question of whether a given structural correspondence is exploited.[11] Firstly, there is a question about the causal efficacy of a given internal relational structure to downstream processing within the LLM itself. Call this the *causal sensitivity question*. Secondly, there is a question about whether LLMs' success at tasks depends on the degree of structural correspondence between this internal structure and some target entities. Call this the *success relevance question*.

#### 4.2.1. Causal sensitivity

The sensitivity condition helps to screen-off cases where an internal structure bears a correspondence to something, but that internal structure has no causal impact on processing or the production of behaviour. To borrow an example from Shea (2014), suppose that neurons differ slightly in colour. There might be a way of mapping neurons in a person's brain to some entities in their environment such that the relative hue of neurons preserves a relation over those entities. But it is not plausible that the relative hue of neurons thereby *represents* that relation. Why? Part of the reason is that it is highly implausible that processing in the brain is causally sensitive to the relative hue of neurons; there is no downstream process that (say) takes the colour difference between two neurons as an input and delivers an output based on that difference (Shea, 2014, pp. 133–4).

---

[10] The suggestion here is not that LLMs memorise their entire training data, exhaustively encoding the frequency of every sequence of tokens. It's that they might rely on internal structures which capture statistical relationships between some important subsets of tokens.
[11] I draw heavily on Shea's (2018) unpacking of the notion of exploitability.

The need to establish causal sensitivity is sometimes emphasised in the AI literature analysing (supposed) representations in LLMs. For example, Meng and colleagues (2022) trace the downstream causal impacts of interventions on particular activations as part of their justification for identifying those activations as vehicles of representational content. Importantly though, this particular study did not explore causal sensitivity to *relations between* activations which is what is at issue when assessing the exploitability of structural correspondences.

Exploitation has also been discussed in some philosophical treatments of representation in LLMs. For instance, "use" is one of the three criteria for representation in Harding's (2023) account of representation in LLMs—where use is operationalised in terms of the causal effects of interventions on candidate representational vehicles. While Harding's account focuses on use or exploitation of *informational* correspondences, Coelho Mollo & Millière (2023) briefly discuss structural correspondences in their broader discussion of representation grounding in LLMs. They too stress that structural correspondences alone are insufficient to ground representational content, though they do not fully unpack what it would take to establish exploitation of a structural correspondence.

Is processing in LLMs causally sensitive to internal structures that stand in structural correspondences to extralinguistic structures? There is some evidence that it is: Merullo, Eickhoff & Pavlick (2024) investigated how transformer-based language models solve a set of tasks such as answering the question "Q: What is the capital of France? A: Paris Q: What is the capital of Poland? A\_\_\_". They showed that the LLMs' answers depended on a vector addition operation performed in certain layers of the network. In these crucial layers, the pattern of activation passed on from the units of the previous layer (which can be conceptualised as a vector) is transformed in a predictable way: a specific vector is added to the initial vector. Following the vector addition operation, the LLM outputs "Warsaw" whereas, without that operation, the model would output

"Poland". Importantly, the effect of the very same vector addition operation seemed to be contextually appropriate across many different inputs. The experimenters demonstrated this by changing the prompt, then intervening on the network with the vector extracted from the "Poland"–>"Warsaw" task. If a vector at a particular layer would otherwise prime the model to produce a country-name as an output, adding the critical vector at that layer resulted in an appropriate capital-name output, largely irrespective of the country-name. For instance, adding it to a "China" vector would result in "Beijing", "Somalia" in "Mogadishu", and so on. Thus, some LLMs seem to rely on a kind of simple arithmetic on vectors (adding one vector to another) when solving certain tasks.

The researchers' interventions were not always successful. Often, the intervention did not result in changing the output, but did result in significantly increasing the reciprocal rank of the target city name (the rank assigned to it in the probability distribution over all tokens, extracted from that layer). And for 37% of cases, the intervention had no effect (this error rate was lower for other relational tasks, e.g. turning verbs into their past tense form).[12] So the underlying mechanisms here are clearly complex and in need of further empirical investigation. Nevertheless, the study hints at the sort of evidence relevant to assessing whether processing in LLMs is sensitive to a certain relation between vectors.

To flesh out this thought, let's somewhat speculatively extend the mechanism proposed by Merullo and colleagues: suppose that processing in certain layers of LLMs computes the offset between the point in activation space assigned to "France" and the point in activation space assigned to "Paris" and then offsets the point in activation space assigned to "Poland" in the same way, to arrive at a point in activation space which produces the output "Warsaw". If this is correct, then at least some relations between vectors are not like relations of relative hue between

---

[12] The authors discuss some possible explanations for these results (Merullo, Eickhoff & Pavlick 2024, p. 6),

neurons—they have a real causal impact on downstream computations, and ultimately the output of the LLM, so they meet the sensitivity constraint on exploitable structural correspondences.[13]

This is just one illustrative example of the kind of data which could be relevant to answering the sensitivity question. Importantly, however, given that many different structural correspondences have been claimed to ground representational content in LLMs, the sensitivity question needs to be answered on a case-by-case basis. Processing in LLMs might be causally sensitive to specific vector arithmetical relations, but are they sensitive to *distance in activation space more broadly*? What about *mereological relations* in activation space (such as a certain region's being a subset of another)? More empirical work is needed to answer these questions.

### 4.2.2. Success relevance

A structural correspondence that is a candidate for grounding representational content involves a relation between two structures (those structures themselves constituted by relations among some entities). On the one side there is an internal structure (e.g. a relation among a set of neural network properties). On the other side there is an external structure (e.g. a relation among a set of physical locations). In the last sub-section, in discussing causal sensitivity to the relations comprising some internal structure, we considered only one side of this equation. Even having established causal sensitivity to some internal structure, a question remains whether that causal sensitivity functions to allow the system to deal with, navigate or respond to the structure on the other end of the candidate structural correspondence.

The success relevance question is the question of whether a system's success at performing tasks depends on the degree of structural correspondence between some internal structure and some external conditions. An internal structure, even one that internal processing is causally sensitive to, might correspond to any number of things, but only some of these correspondences will be explanatorily relevant to the whole system's (e.g. the organism's or the LLM's) successful behaviour. To give an exaggerated example, distances among points in the activation space of neurons in a frog's brain might, through sheer chance, bear a structural correspondence to distances among asteroids in a distant galaxy. But what goes on in distant galaxies is none of the frog's business (so to speak), and so this structural correspondence is unlikely to be explanatory of the frog's behavioural success, even if processing in the frog's brain is causally sensitive to distance in neural activation space.

---

[13] It is a fair question how this comparison operation could occur (I thank an anonymous reviewer for raising this). Presumably the self-attention layers play some role, as they take two vectors as input when moving information between token positions. However, it's also plausible that the offset of these vectors is calculated in the subsequent multi-layer perceptron (MLP) block, by some transformation that disentangles the contribution of the attention layer from the original vector for that token position. Determining whether and, if so, how models are causally sensitive to offsets in embedding space is a job for future empirical research.

Yildirim & Paul (2024) seem to have something like this in mind when they say that, to qualify as a world-model, a structure in an LLM "must be behaviorally efficacious, meaning that it should enable *accurate* planning and *high-reward* actions back in the real world" (p. 2, emphasis added). Evidently, answering the success relevance question for LLMs is going to depend on *extrinsic* facts about LLMs—on the kind of environment they are embedded in and on what tasks they are called upon to perform. These tasks introduce success conditions, and thus constrain the pool of structural correspondences that are candidates for being explanatory (being causally relevant to achieving successful outcomes).

How does this bear on the possibility of LLM's representing real world (as opposed to linguistic) entities? If the tasks which a system performs constrain what it is able to represent, then a version of the worry raised by Bender and Koller's octopus thought experiment presents itself again: text-bound LLMs don't perform tasks directly involving real world entities—their inputs and outputs consist solely of text—and so this might naturally be thought to prevent them from representing real world domains (regardless of whether their internal states happen to structurally correspond to those domains). Of course, one could modify a text-bound LLM such that their capacities did involve acting on or responding to real world entities. Multimodal and embodied

LLMs provide examples.[14] But my focus in this paper is on *text-bound* LLMs, which have no text-unmediated access to the real world.[15]

Despite this challenge, I will show that the text-bound nature of LLMs does not immediately rule out real world contents on a structural correspondence account of representation: in principle, the degree of correspondence between LLMs' internal structures and real world entities might still explain their success on text-based tasks. Deciding how best to characterise those tasks, however, turns out to be a non-trivial issue.

### 4.2.3. What tasks do LLMs perform?

LLMs' basic functionality involves turning text inputs into text outputs. But how should we think about these capacities? They could be described as, say, summarising, translating, answering factual questions, offering recommendations. These descriptions, and the success conditions they imply, involve treating the inputs and outputs as meaningful (i.e. individuating them semantically), and for certain tasks, comparing those meanings to some external, real world standard (the facts of the matter). For example, a good (successful) answer to a factual question ("Paris") is one whose content is appropriate given the content of the input ("What is the capital of France?"), and the way the world in fact is (the fact that Paris is the capital of France). But, for any given capacity, there will be a competing description of that capacity in non-semantic terms. For example, we could describe an LLM as engaged in the task of predicting the most statistically likely token to follow a sequence of tokens (for a wide range of input sequences). Here the inputs and outputs are individuated in purely formal terms, and success is a matter of tracking the actual statistics of human generated language (rather than tracking any extra-linguistic facts).

---

[14] SayCan (Ahn et al., 2022) is an implementation of an LLM in an instruction-following robot. Receiving the prompt "I spilled my drink, can you help?" the system can break down this request into contextually appropriate sub-tasks and then execute these tasks in the real world (locating a sponge, picking it up, locating the spillage, using the sponge to clean it up). An interesting intermediate case is a system that can take actions in a virtual environment like a web-browser, such as Adept's ACT-1 Model (Adept 2022).

[15] This distinguishes the challenge addressed in this paper from applications of structural representation based accounts of representation to artificial neural networks in previous philosophical work (Churchland 1998; 2012; O'Brien & Opie, 2004, 2006).

These different ways of describing the system's capacities are significant because they result in different accounts of what success constitutes, and what is successful relative to one task might be unsuccessful relative to another. For example, consider the study by Merullo and colleagues described above in Section 4.2.1. If we think of the task that the LLM performs as the task of correctly identifying the capital of a country, then successful performance will require that when asked about the capital of Poland, the model answers "Warsaw". But now suppose that the capital city of Poland officially changes from Warsaw to Kraków, but the model is not retrained, such that it continues to answer "Warsaw". For the sake of a clean example, let's also suppose that, for a short while, the news about the change to Poland's capital is not widely publicised, and most people continue to (now incorrectly) finish the sentence "the capital city of Poland is ____" with "Warsaw". Like those humans, the LLM's answer "Warsaw", would (given the new facts) count as an *unsuccessful* performance of the task, on this construal of the task that the model is engaged in. By contrast, if the model's task is just predicting the most statistically likely next word or token, its answer "Warsaw" remains successful even after the change in the geopolitical facts. This is because (by assumption) the statistics of human-produced language are largely unchanged, and "Warsaw" remains the most statistically likely answer to that question.

If the question of whether a structural correspondence is exploited depends on whether it contributes to the system's successful performance of its tasks, then answering this question requires first deciding between these competing accounts of its tasks (and the success conditions they imply). How can we do this?

### 4.2.4. Training history as a determinant of success conditions

In the case of biological organisms, a prominent view is that the success conditions for an organism's behaviour derive from what is "good for" the system or, relatedly, about how behaviours are rewarded or reinforced, broadly in keeping with the teleosemantic tradition (Dretske 1988; Millikan 1984). While there are different accounts on offer (Maley & Piccinini, 2017; Piccinini, 2022; see Lee 2021 pp. 822–823 for discussion), one promising approach takes a backward-looking view, holding that success conditions are determined by the factors which played a causal role in stabilising behavioural dispositions in the history of the organism, evolutionary lineage, or system (Shea 2018).

Shea (2018) discusses three kinds of stabilising processes which can play this role: (i) evolution by natural selection, (ii) contribution to the persistence of a single organism or system and (iii) learning based on feedback. In each case, there is some process that determines which dispositions survive and persist into the future (either in a single organism or in its descendants). For instance, natural selection will tend to stabilise, in a population, dispositions to acquire food, as individuals who lack this disposition are less likely to proliferate their genes (evolution by natural selection). And, at the timescale of a single individual, an organism and its dispositions will survive for longer if one of those dispositions is to acquire food (contribution to the persistence of a single organism or system). Learning mechanisms play a similar role, typically on shorter timescales: a mechanism that reinforces behaviours when they lead to high value outcomes will tend to stabilise certain dispositions, such as a disposition that results in acquiring food, over others (learning based on feedback).

The existence of these processes provides a non-arbitrary way to distinguish between successful outcomes (to speak metaphorically, outcomes that an organism's behaviour is "trying to" or "supposed to" achieve) and accidental or unsuccessful outcomes. Successful outcomes are those which have been the target of stabilising processes; unsuccessful or accidental outcomes are those which have not. That is, we call an outcome of behaviour on a particular occasion successful if that outcome is of a kind that, in the history of the organism, helped to create or reinforce the disposition to produce that very behaviour.

Importantly, this way of looking at things allows us to get reasonably fine grained in specifying the tasks an organism is performing. A frog might have a robust disposition to snap with its tongue at little black flying things, but it is the fact that little black flying things have tended to be *flies* (or *nutritious objects*) that has stabilised frogs' disposition to snap at them with their tongues. Thus, the latter properties, but not the former, play a role in a causal explanation of the disposition's stabilisation, and thus provide the correct characterisation of the task the frog is performing (and its attendant success conditions—it is unsuccessful when it captures a little black flying thing that is not a fly, or not nutritious) (Shea 2018, pp. 150–151).

While the account just outlined was developed primarily in the context of biological organisms, it suggests a parallel strategy for determining the success conditions for an LLM's dispositions: we need to look at the processes which lead to those dispositions being stabilised and maintained. And the natural place to look for this is the *machine learning processes* through which LLMs are trained.

Presently, most LLMs are initially trained by a self-supervised machine learning process on the task of *next token prediction*. A "token" refers to words or sub-word parts that are treated as the basic units of text strings by the LLM. (For example, "statistical relationship" might be broken into three tokens: "stat", "istical" and "relationship"). This first stage of training for an LLM, often referred to as "pre-training", involves giving the network a portion of a string of text from its training corpus (e.g. an incomplete paragraph, or a sentence with one word masked). The network's output is a probability distribution over all possible next tokens. The highest probability constitutes its "best guess" at the next token in the sequence, and is used to generate the output. The actual next word is then revealed and, if the initial guess was incorrect, the learning algorithm updates the connection weights in the model that led to the incorrect answer. The process is repeated iteratively and, through a process of gradient descent, the model gets better at the task of next-word prediction.

Some LLMs are trained simply on this self-supervised next word prediction task ("pre-training"). Other LLMs undergo a further training process ("fine tuning"). For example, OpenAI's LLMs (since GPT-3.5) undergo a process of reinforcement learning from human feedback (RLHF) (Ouyang et al., 2022). Pairs of responses to a sample of prompts are shown to human reviewers, and data is collected on the reviewers' relative preferences for responses to each prompt. This dataset is then used to train a distinct neural network—a reward model (or "preference model"). The reward model takes prompt–response pairs as input and generates a scalar rating as output. Once trained, the reward model acts as a proxy for human reviewers and thus its output can be used as reward signal to fine-tune the original pre-trained LLM in a reinforcement learning process.

Crucially, the different feedback signals in these two training runs motivate different interpretations of the tasks an LLM is engaged in.[16] In pre-training, the measure of the model's performance (the "loss function" or "cost function") that is used as feedback to update the model's trainable parameters, is based on the difference between the actual next token (in the training example), and the predicted next token (the model's output). Thus, pre-training stabilises a disposition to produce statistically probable continuations of a sequence of words. Statistically probable continuations of a sequence of words certainly *correlate* with semantically appropriate responses (at least in a training dataset of sufficient size and quality). But, in self-supervised pre-training, the causally relevant factor driving the updating of the models' trainable parameters, and thus its dispositions, is the probability of word-occurrences conditional on the previous string of words. If, for some reason, the most statistically likely continuation of a particular sequence is semantically inappropriate (e.g. false, incoherent, or inconsistent) the training procedure would not register this, so the disposition to produce such an output wouldn't have been selected against.

---

[16] Coelho Mollo and Millière (2023) argue for a similar conclusion.

By contrast, fine-tuning with RLHF explicitly involves selecting answers based on semantically laden criteria. For example, human evaluators may be asked to rate how "helpful, honest, and harmless" pretrained LLMs' outputs are (Bai et al., 2022), these evaluations then being fed into the reward signal used to fine-tune the model. Here, the training process does seem to stabilise the model's outputs towards semantically appropriate outputs. Unlike pre-training, fine tuning tends to select against semantically inappropriate outputs *in virtue of their semantic inappropriateness*. The semantic (in)appropriateness of its outputs causally drives the stabilisation of its dispositions.

Thus, according to a stabilisation-history dependent account of the tasks in which LLMs are engaged, LLMs trained in different ways are properly seen as engaging in different whole-system-level tasks. While purely pretrained LLMs have the function or goal of predicting the most probable next token, those fine-tuned by RLHF and similar processes can develop tasks that involve producing semantically appropriate (including truth-tracking) responses to prompts. This is because the semantic appropriateness of their outputs is not merely a fortunate side effect, but played a historical causal role (via a human-mediated training algorithm) in their having the dispositions that they now have.

### 4.2.5. Do structural correspondences to real world structures contribute to LLMs' successful task performance?

Consider a pre-trained LLM whose system-level task is simply to generate probable next tokens. It's conceivable that a system of this kind achieves its goal partly by exploiting structural correspondences between its internal activation spaces and real world structures. We can, for instance, conceive of a model that uses a representation of the structural facts in some worldly domain as an intermediate step on the way to calculating likely continuations of text.[17] For example, if the input context suggests that the model should output text in the style of a person with poor geographic knowledge, such a model might start from a representation of the actual capital–country relations and then modulate its answer away from the truth, to generate a statistically probable output.

But it's also possible that a pre-trained LLM does not involve any such mechanism. It might rely directly on structural correspondences to statistical co-dependence relations between one class of words ("China", "France", "Poland" etc.) and another ("Beijing", "Paris", "Warsaw" etc.), without going via an intermediate computational step whose function is to correspond to real world country–capital relations. Thus, simply identifying an LLM's system-level task does not settle the question of which of the competing structural correspondences (involving real world structures, vs. purely linguistic-statistical structures) is exploited.

How can we tease apart these two hypotheses empirically? One way of evaluating which structural correspondence is explanatory of success is by intervening on the system to modulate the degree of each structural correspondence.[18] If there are multiple correspondences competing for explanatory relevance, these could be manipulated independently to see which has the greater effect on success (relative to the success criteria applicable to that system's outputs). That is, one could artificially adjust the geometry of an LLM's activation space at a layer of interest, so as to differentially tighten the structural correspondence to either (i) the real world structure or (ii) the linguistic-statistical structure.[19] If the first hypothesis is true—the LLM relies on a step that exploits a correspondence to actual country–capital relations—we should find a layer in which increasing the correspondence with (i) has a greater positive effect on performance than increasing the correspondence with (ii) (all else being equal).[20] "Positive effect" here means generating more *probable* tokens at output across the range of relevant contexts, given that in this case, this is the metric against which we measure the system's success.[21]

---

[17] I thank two anonymous reviewers for pressing me to address this kind of possibility.
[18] Shea (2018, pp. 142–143) discusses the evidential role of interventions on structural correspondences.
[19] For such an experiment to be possible, it must be the case that (i) and (ii) are not themselves in perfect correspondence, so that they can be differentially modulated. This is plausible: it is likely that there are many cases where the more common co-occurrences are between word pairs, such as "Australia"–"Sydney" and "Nigeria"–"Lagos", that diverge from country–capital pairs.

The same test could be performed on a system that has been fine-tuned with RLHF (and thus has system-level tasks that involve generating truth-tracking outputs). But, for these systems (in contrast to purely pre-trained systems) we would need to measure the effect of modulating a correspondence not in terms of whether more probable answers are generated, but whether more semantically appropriate, truth-tracking outputs are generated. In principle, the different success criteria could yield a different verdict on which correspondence is exploited.

Above, I noted that it is conceivable that a purely pretrained LLM develops mechanisms that function to correspond to real world structures, as a means of generating statistically probable outputs. But it bears noting that in the case of systems fine-tuned with RLHF, there is additional reason to expect them to develop internal structures whose function is to structurally correspond to real world structures, given that a key task of such systems is to generate outputs that respect real world facts.

Some recent empirical work points toward the sort of intervention experiment needed here. Chen and colleagues (2023) built on Gurnee and Tegmark's investigation of spatial correspondences in LLMs (discussed above in Section 4.1). They sought to determine whether structural correspondences with spatial structures were causally relevant to performance. Like Gurnee and Tegmark, they trained probing regressors on the internal activations of two LLMs (in this case DeBERTa-v2 and GPT-Neo). The probe's objective was to infer from the internal activation the latitudes and longitudes corresponding to the location named in the model's current

---

[20] When *weakening* the correspondences, we should see a greater *negative* impact for the exploited correspondence.
[21] Harding (2023) suggests a similar methodology, though her focus is informational correspondences rather than structural correspondences. She argues that we can determine whether an informational correspondence is exploited by intervening on the internal state and seeing if it "degrades performance" relative to some "goodness measure" (pp. 13–14). I differ from Harding in that she is content to "stipulat[e] the task … (rather than searching for a world-involving task that is independent of human interpretation)" (p. 9).

input. As in the previous study, their probes identified internal gradients in the models' activation spaces which were taken to be candidates for encoding spatial dimensions of variation (latitude and longitude). To assess the causal relevance of these identified gradients on the models' performance, the researchers perturbed the model's internal activations, pushing them along that gradient either closer to or further away from the "correct" activations, as determined by the probe. For example, if an input mentioned Tokyo, the intervention would bring the model's internal activation closer to or further away from the activation that the probe implied it should have, given the actual coordinates of Tokyo. Thus, in effect, they brought a set of (putative) representational vehicles into stronger or weaker structural correspondence with a real world structure. The researchers found that perturbing the activations in the opposite direction from the (presumed) correct activations had a strong and statistically significant negative impact on the model's ability to give correct answers to prompts of the form "<city name> is located in the country of…"[22]

While this study gives a sense of the kind of interventions I have in mind, two things bear noting here. Firstly, the two models used by the authors are pretrained on next-token-prediction style objectives, with no reinforcement learning from human feedback. As I argued in the previous sub-section, this means that their success should not be measured against the *ground truth* answers to geographical questions—which was how Chen et al. (2023) assessed accuracy—but relative to the *probability* of a given answer. Secondly, Chen and colleagues only evaluated interventions on the correspondence with the real world spatial structure. To firmly establish that this was the exploited correspondence, one would need to compare the magnitude of these effects to other candidates for exploited correspondences, which may include co-occurrence structures over classes of words.

---

[22] Perturbing the activations *towards* the correct activations saw a much weaker (though still statistically significant) positive improvement on performance. The authors conjecture that "the LLM was already close to its optimal capacity, so the augmented spatial information in the activation after gradient descent didn't help much with the prediction" (p. 10).

## 5. Conclusion

A key question in interpreting LLMs is whether we should understand text-bound versions of these models as representing real world entities and structures. In applying a structural correspondence based account of representation, we saw that the mere existence of structural correspondences between LLM's internal states and real world structures is insufficient to ground representation of those real world structures—those correspondences must be exploited. Despite some prima facie challenges, I showed that the text-boundedness of LLMs is not an in-principle barrier to an LLM exploiting such correspondences. While we saw some suggestive empirical evidence, determining whether (and for which real world structures) this in principle possibility is in fact realised requires further empirical investigation: Firstly, more work is required to identify which model-internal structures processing in LLMs is causally sensitive to. Secondly, to decide between alternative mechanistic hypotheses, work is required to compare the effects on success of modulating different structural correspondences. As I have argued, such work should be careful to apply appropriate success criteria, given the training history of the LLM in question.

**References**

Abdou, M., Kulmizev, A., Hershcovich, D., Frank, S., Pavlick, E., & Søgaard, A. (2021). Can Language Models Encode Perceptual Structure Without Grounding? A Case Study in Color. *Proceedings of the 25th Conference on Computational Natural Language Learning*.

Ahn, M., Brohan, A., Brown, N., Chebotar, Y., Cortes, O., David, B., … Zeng, A. (2022). *Do As I Can, Not As I Say: Grounding Language in Robotic Affordances*. arXiv Preprint. https://arxiv.org/abs/2204.01691

Bai, Y., Jones, A., Ndousse, K., Askell, A., Chen, A., DasSarma, N., … Kaplan, J. (2022). *Training a Helpful and Harmless Assistant with Reinforcement Learning from Human Feedback*. arXiv Preprint. http://arxiv.org/abs/2204.05862

Belinkov, Y. (2022). Probing Classifiers: Promises, Shortcomings, and Advances. *Computational Linguistics*, *48*(1), 207–219. https://doi.org/10.1162/COLI_a_00422

Bender, E. M., & Koller, A. (2020). Climbing towards NLU: On Meaning, Form, and Understanding in the Age of Data. *Proceedings of the 58th Annual Meeting of the Association for Computational Linguistics*, 5185–5198.

Butlin, P. (2023). Sharing Our Concepts with Machines. *Erkenntnis*, (88), 3079–3095. https://doi.org/10.1007/s10670-021-00491-w

Chen, Y., Gan, Y., Li, S., Yao, L., & Zhao, X. (2023). *More than Correlation: Do Large Language Models Learn Causal Representations of Space?* arXiv Preprint. https://arxiv.org/abs/2312.16257

Churchland, P. M. (1998). Conceptual similarity across sensory and neural diversity: The Fodor/Lepore challenge answered. *Journal of Philosophy*, *95*, 5–32.

Churchland, P. M. (2012). *Plato's camera: How the physical brain captures a landscape of abstract universals*. Cambridge, MA: MIT Press.

Coelho Mollo, D., & Millière, R. (2023). *The Vector Grounding Problem*. 1–34. arXiv Preprint. http://arxiv.org/abs/2304.01481

Corneil, D., & Gerstner, W. (2015). Attractor network dynamics enable preplay and rapid path planning in maze-like environments. *Advances in Neural Information Processing Systems*, 1684–1692.

Cummins, R. (1996). *Representations, Targets, and Attitudes*. Cambridge, MA: MIT Press.

Dretske, F. (1988). *Explaining Behavior*. Cambridge, MA: MIT Press.

Geva, M., Schuster, R., Berant, J., & Levy, O. (2021). Transformer Feed-Forward Layers Are Key-Value Memories. *Proceedings of the 2021 Conference on Empirical Methods in Natural Language Processing,* 5484–5495. https://doi.org/10.18653/v1/2021.emnlp-main.446

Gładziejewski, P. (2015). *Explaining cognitive phenomena with internal representations: a mechanistic perspective. 40*(53), 63–90. https://doi.org/10.1515/slgr-2015-0004

Gładziejewski, P. (2016). Action guidance is not enough, representations need correspondence too: A plea for a two-factor theory of representation. *New Ideas in Psychology*, *40*, 13–25. https://doi.org/10.1016/j.newideapsych.2015.01.005

Gładziejewski, P., & Miłkowski, M. (2017). Structural representations: causally relevant and different from detectors. *Biology and Philosophy*, *32*(3), 337–355. https://doi.org/10.1007/s10539-017-9562-6

Godfrey-Smith, P. (1996). Complexity and the Function of Mind in Nature. In *Complexity and the Function of Mind in Nature*. https://doi.org/10.1017/cbo9781139172714.007

Godfrey-Smith, P. (2006). Mental Representation, Naturalism, and Teleosemantics. In D. Papineau & G. Macdonald (Eds.), *Teleosemantics*. Oxford: Oxford University Press.

Gurnee, W., & Tegmark, M. (2024). Language Models Represent Space and Time. *The Twelfth International Conference on Learning Representations*. https://arxiv.org/abs/2310.02207

Harding, J. (2023). Operationalising Representation in Natural Language Processing. *The British Journal for the Philosophy of Science*, 1–36. Advanced Online Publication. https://doi.org/10.1086/728685

Hernandez, E., Li, B. Z., & Andreas, J. (2024). Inspecting and Editing Knowledge

Representations in Language Models. *Conference on Language Modelling*.

https://arxiv.org/abs/2304.00740

Isaac, A. M. C. (2013). Objective similarity and mental representation. *Australasian Journal of Philosophy*, *91*(4), 683–704. https://doi.org/10.1080/00048402.2012.728233

Khajeh-Alijani, A., Urbanczik, R., & Senn, W. (2015). Scale-free navigational planning by neuronal traveling waves. *PLoS ONE*, *10*(7), 1–15. https://doi.org/10.1371/journal.pone.0127269

Kosinski, M. (2024). *Evaluating Large Language Models in Theory of Mind Tasks. Proceedings of the National Academy of Sciences*, *121*(45), e2405460121.

Kriegeskorte, N., Mur, M., & Bandettini, P. (2008). Representational similarity analysis - connecting the branches of systems neuroscience. *Frontiers in Systems Neuroscience*, *2*(NOV), 1–28. https://doi.org/10.3389/neuro.06.004.2008

Lee, J. (2019). Structural Representation and the two problems of content. *Mind & Language*, *34*(5), 606-626. https://doi.org/10/gfkmb5

Lee, J. (2021). Rise of the swamp creatures: Reflections on a mechanistic approach to content. *Philosophical Psychology*, *34*(6), 1–24. https://doi.org/10.1080/09515089.2021.1918658

Lee, J., & Calder, D. (2023). The many problems with S-representation (and how to solve them). *Philosophy and the Mind Sciences*, *4*, 8. https://doi.org/https://doi.org/10.33735/phimisci.2023.9758

Maley, C. J., & Piccinini, G. (2017). A Unified Mechanistic Account of Teleological Functions for Psychology and Neuroscience. In D. M. Kaplan (Ed.), *Explanation and Integration in Mind and Brain Science* (pp. 236–256). Oxford: Oxford University Press.

Meng, K., Bau, D., Andonian, A., & Belinkov, Y. (2022). Locating and Editing Factual Associations in GPT. *Advances in Neural Information Processing Systems*, *35*(NeurIPS).

Millikan, R. (1984). *Language, Thought, and Other Biological Categories*. Cambridge, MA: MIT Press.

Mikolov, T., Yih, W., & Zweig, G. (2013). Linguistic Regularities in Continuous Space Word

Representations. *Proceedings of the 2013 Conference of the North American Chapter of the Association for Computational Linguistics: Human Language Technologies*, 746–751.

Mitchell, M., & Krakauer, D. C. (2022). The Debate Over Understanding in AI's Large Language Models. *Proceedings of the National Academy of Sciences*, *120*(13), e2215907120.

Newman, M. H. A. (1928). *Mr. Russell's 'Causal Theory of Perception.' Mind*, *XXXVII*(146), 137–148. https://doi.org/10.1093/mind/XXXVII.146.137

Nirshberg, G. (2023). Structural Resemblance and the Causal Role of Content. *Erkenntnis*, (0123456789). https://doi.org/10.1007/s10670-023-00699-y

O'Brien, G., & Opie, J. (2004). Notes toward a structuralist theory of mental representation. In H. Clapin, P. Staines, & P. Slezak (Eds.), *Representation in mind: New approaches to mental representation* (pp. 1–20). https://doi.org/10.1016/B978-008044394-2/50004-X

O'Brien, G., & Opie, J. (2006). How do connectionist networks compute? *Cognitive Processing*, *7*(1), 30–41. https://doi.org/10.1007/s10339-005-0017-7

O'Keefe, J., & Burgess, N. (1996). Geometric determinants of the place fields of hippocampal neurons. Nature, 381(6581), 425–428.

O'Keefe, J. & Nadel, L. (1978). *The Hippocampus as a Cognitive Map*. Oxford: Clarendon

Ouyang, L., Wu, J., Jiang, X., Almeida, D., Wainwright, C. L., Mishkin, P., … Lowe, R. (2022). *Training language models to follow instructions with human feedback*. Retrieved from http://arxiv.org/abs/2203.02155

Park, S. A., Miller, D. S., & Boorman, E. D. (2021). Inferences on a multidimensional social hierarchy use a grid-like code. *Nature Neuroscience*, *24*(9), 1292–1301. https://doi.org/10.1038/s41593-021-00916-3

Pavlick, E. (2023). Symbols and grounding in large language models. *Philosophical Transactions of the Royal Society A*, *381*.

Piccinini, G. (2022). Situated Neural Representations: Solving the Problems of Content. *Frontiers in Neurorobotics*, *16*, 1–13. https://doi.org/10.3389/fnbot.2022.846979

Ramsey, W. M. (2007). *Representation reconsidered*. Cambridge: Cambridge University Press. https://doi.org/10.1017/CBO9780511597954

Reid, A. K., & Staddon, J. E. R. (1998). A Dynamic Route Finder for the Cognitive Map. *Psychological Review*, *105*(3), 585–601. https://doi.org/10.1037/0033-295X.105.3.585

Rogers, A., Kovaleva, O., & Rumshisky, A. (2020). A primer in BERTology: What we know about how BERT works. *Transactions of the Association for Computational Linguistics*, *8*, 842–866. https://doi.org/10.1162/tacl_a_00349

Searle, J. R. (1980). Minds, brains, and programs, *Behavioral and Brain Sciences*, *3*, 417–457.

Shea, N. (2013). Millikan's Isomorphism Requirement. In D. Ryder, J. Kingsbury, & K. Williford (Eds.), *Millikan and Her Critics* (pp. 63–80). Oxford and Malden, MA: Wiley- Blackwell.

Shea, N. (2014). Exploitable isomorphism and structural representation. *Proceedings of the Aristotelean Society*, *114*(2), 123–144. https://doi.org/10.1111/j.1467-9264.2014.00367.x

Shea, N. (2018). *Representation in Cognitive Science*. Oxford: Oxford University Press.

Søgaard, A. (2022). Understanding models understanding language. *Synthese*, *200*(6), 1–16. https://doi.org/10.1007/s11229-022-03931-4

Søgaard, A. (2023). Grounding the Vector Space of an Octopus: Word Meaning from Raw Text. *Minds and Machines*, *33*(1), 33–54. https://doi.org/10.1007/s11023-023-09622-4

Swoyer, C. (1991). Structural Representation and Surrogative Reasoning. *Synthese*, *87*, 449–508.

Tenenbaum, J. B., Kemp, C., Griffiths, T. L., & Goodman, N. D. (2011). How to grow a mind: Statistics, structure, and abstraction. *Science*, *331*(6022), 1279–1285. https://doi.org/10.1126/science.1192788

Ullman, T. (2023). *Large Language Models Fail on Trivial Alterations to Theory-of-Mind Tasks*. Retrieved from http://arxiv.org/abs/2302.08399

Williams, D., & Colling, L. (2018). From symbols to icons: the return of resemblance in the cognitive neuroscience revolution. *Synthese*, *195*(5), 1941–1967. https://doi.org/10.1007/s11229-017-1578-6

Yildirim, I., & Paul, L. A. (2024). From task structures to world models: what do LLMs know ? *Trends in Cognitive Sciences*, 1–12. https://doi.org/10.1016/j.tics.2024.02.008